\PassOptionsToPackage{table}{xcolor}
\documentclass{article} % For LaTeX2e
\usepackage{seekvln_preprint,times}

\usepackage{amsmath,amsfonts,bm}

\def\eqref#1{equation~\ref{#1}}
\def\1{\bm{1}}

\DeclareMathAlphabet{\mathsfit}{\encodingdefault}{\sfdefault}{m}{sl}
\SetMathAlphabet{\mathsfit}{bold}{\encodingdefault}{\sfdefault}{bx}{n}

\usepackage{hyperref}
\hypersetup{hidelinks}
\usepackage{url}

\usepackage{graphicx} % for \includegraphics, \resizebox, ...
\usepackage{xcolor} % for \rowcolor, cell colors in tables

\definecolor{myviolet}{HTML}{800080}

\makeatletter
\newcommand{\appendixcontents}{\section*{Appendix Contents}\@starttoc{apc}}
\makeatother
\newcommand{\appendixsectionentry}[1]{\addcontentsline{apc}{section}{\protect\numberline{\thesection}#1}}
\newcommand{\appendixsubsectionentry}[1]{\addcontentsline{apc}{subsection}{\protect\numberline{\thesubsection}#1}}

\usepackage{booktabs}
\usepackage{longtable} % allow the RL configuration table to continue across pages
\usepackage{multirow}
\usepackage{amssymb}
\usepackage{caption} % for independent captions inside side-by-side minipages
\usepackage{float} % for placing paired analysis tables beside their discussion
\usepackage{needspace} % keep inline headings with their opening text
\usepackage{mdframed} % framed prompt templates in the appendix
\usepackage{fvextra} % line-wrapping verbatim text inside prompt boxes
\usepackage[T1,OT1]{fontenc} % T1 for prompt symbols; keep OT1 as the manuscript default
\newmdenv[
  linewidth=0.6pt,
  linecolor=black!55,
  backgroundcolor=black!3,
  skipabove=6pt,
  skipbelow=6pt,
  innerleftmargin=7pt,
  innerrightmargin=7pt,
  innertopmargin=6pt,
  innerbottommargin=6pt
]{promptbox}

\title{Seek Before You Move: Evidence Seeking for Progress Grounding in Vision-Language Navigation}

\author{\begin{minipage}{0.96\textwidth}
\centering
Zhimin Wang$^{1,2}$, Meiyuan Zhu$^{1}$, Duo Wu$^{1,2}$, Linjia Kang$^{1}$\\
Yajun Wang$^{3,4}$, Yuan Ni$^{5}$, Xiaohang Wang$^{5}$, Tianlu Pan$^{2}$\\
Jingyan Jiang$^{1}$, Yaowei Wang$^{6,2,\dagger}$, Zhi Wang$^{1,\dagger}$\\[0.6ex]
{\normalfont\small
$^{1}$Tsinghua University \quad $^{2}$Pengcheng Laboratory\\
$^{3}$South China University of Technology \quad $^{4}$International Digital Economy Academy\\
$^{5}$Ping An Technology (Shenzhen) Co., Ltd., Shenzhen, China\\
$^{6}$Harbin Institute of Technology (Shenzhen)\\
$^{\dagger}$Corresponding authors}
\end{minipage}}

\iclrfinalcopy % Show the author block and remove the review ruler.

\begin{document}

\maketitle

\begin{abstract}
Vision-Language Navigation (VLN) requires agents to continuously ground task progress from long-horizon instructions and partial egocentric observations. %Recent VLM-based navigation agents have substantially advanced visual understanding and task planning. Despite these advances, 
Existing VLM-based navigation agents typically reason only over available observations and may remain confident even when task-relevant evidence is missing. For example, an agent may confidently proceed forward and get lost even though the landmark indicating the next turn lies outside its current field of view. We term this failure mode \textbf{\emph{Progress Myopia}}: the agent fails to recognize unreliable progress grounding and continues acting on insufficient evidence. To address it, we propose \textbf{SeekVLN}, an evidence-seeking framework that couples semantic progress reasoning with active acquisition of task-relevant observations. SeekVLN is trained in two stages: First, \textbf{Future-guided Reverse Generation (FRG)} uses future expert actions to augment offline expert trajectories with supplementary views and evidence annotations. Supervised fine-tuning on these trajectories establishes a prior for evidence seeking and progress reasoning without additional expert interaction. However, imitation alone does not reveal whether seeking improves subsequent navigation. We therefore introduce \textbf{Counterfactual} \textbf{Contrastive} \textbf{Policy} \textbf{Optimization} \textbf{(C2PO)} for reinforcement fine-tuning. By comparing each evidence-seeking branch with a counterfactual direct-navigation branch from the same state, C2PO uses a contrastive reward to assign credit to seeking decisions based on subsequent navigation benefit. Experiments on simulated benchmarks show that SeekVLN achieves state-of-the-art performance, improving success rate by 12.7\% and 7.5\% over the base model on R2R-CE and RxR-CE, respectively. Both simulated and real-world evaluations exhibit human-like evidence-seeking behaviors for more reliable progress grounding.
\end{abstract}

\section{Introduction}

Vision-Language navigation (VLN) requires an embodied agent to follow natural-language instructions over long horizons and reach a specified goal~\citep{anderson2018vision,gu2022vision,wu2024vision}. Despite substantial progress~\citep{hong2021vln,an2024etpnav,zhang2024navid}, reliable navigation remains challenging, because success requires continuously grounding the agent's progress: determining completed subgoals and where to go next. In realistic environments, instruction ambiguity and partial egocentric observations often leave the agent with insufficient evidence, introducing substantial uncertainty into progress grounding.

\begin{figure}[t]
    \centering
    \includegraphics[width=\columnwidth]{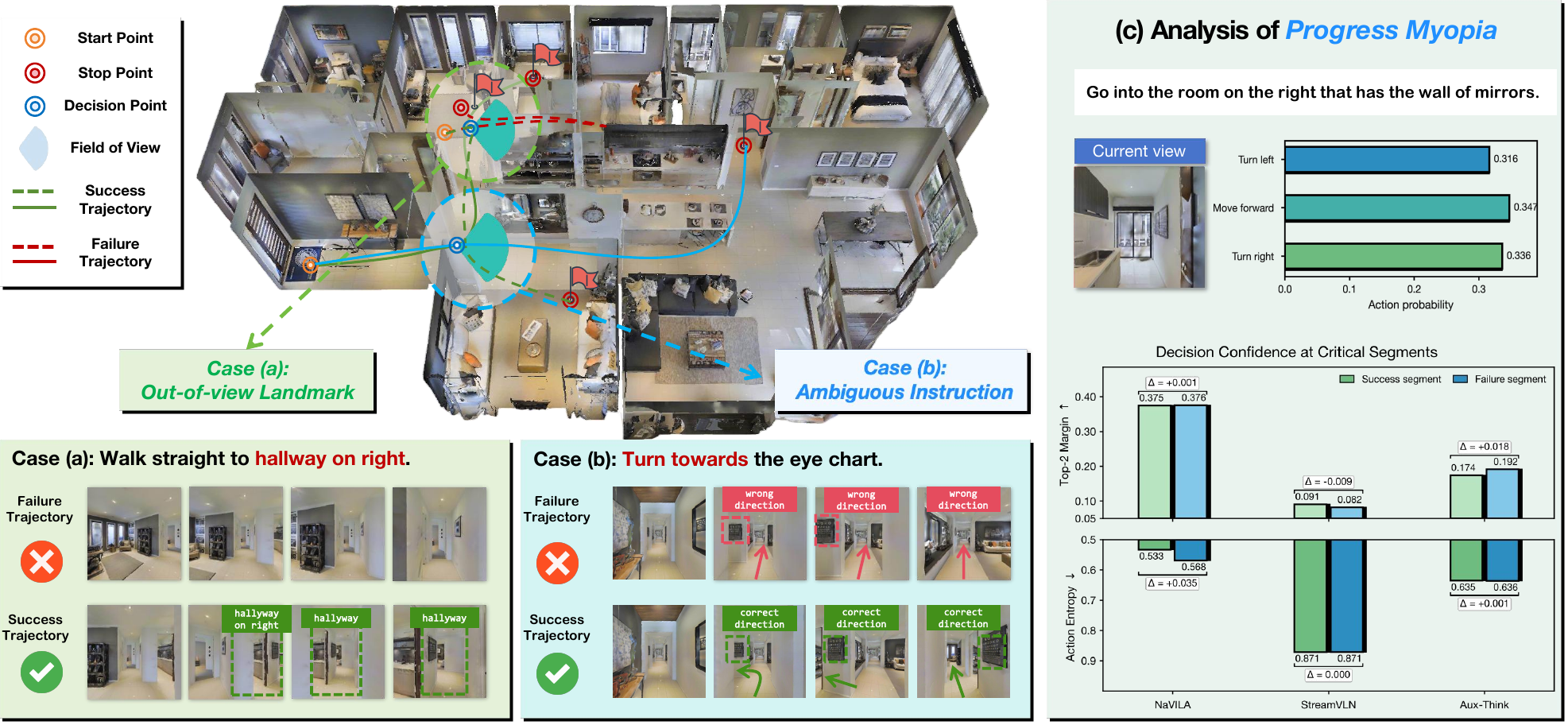}
    \caption{Illustration of \textbf{\emph{Progress Myopia}}. Left: out-of-view landmarks (a) and ambiguous instructions (b) make the current observation insufficient to determine the correct decision. Right: representative navigation agents show similar decision confidence and action entropy on both successful and failed segments, indicating that unreliable decisions can remain highly confident.}
    \label{fig:first-figure}
\end{figure}

As illustrated in Fig.~\ref{fig:first-figure}, progress uncertainty often arises at critical decision points. In case (a), the next landmark falls outside the agent's field of view and therefore cannot be verified from the current observation. At the fork in case (b), neither the instruction nor the current observation provides sufficient evidence to determine the correct direction. These cases illustrate how instruction ambiguity and partial observations introduce uncertainty into progress grounding: the available evidence may be insufficient to determine the agent's current progress and the next navigation action.
%These cases reveal a fundamental cause of navigation uncertainty: the agent may lack the evidence required to determine its current progress and make a reliable decision. 
%These cases indicate that progress uncertainty is not simply caused by limited vision-language understanding but by insufficient evidence for grounding the agent's current progress.

Recent advances in vision-language models (VLMs) have improved visual understanding and long-horizon planning, motivating their use in autonomous navigation agents~\citep{uninavid,navfom,octonav}. Existing agents either directly predict navigation actions from historical context~\citep{navila,streamvln,uninavid}, or improve progress grounding through linguistic reasoning~\citep{progressthink} and landmark-centric representations~\citep{dualanchoring}. However, these methods primarily focus on improving navigation decisions from available observations, with limited attention to assessing the uncertainty in progress grounding.
When task-relevant evidence is insufficient, as in cases (a) and (b) of Fig.~\ref{fig:first-figure}, stronger visual understanding and reasoning do not necessarily enable an agent to recognize that its progress grounding is unreliable.

%However, existing VLM-based VLN methods largely overlook navigation uncertainty: they typically ground task progress solely on the observed trajectory and current egocentric observation. They either directly predict navigation actions from historical context~\citep{navila,streamvln,uninavid}, or improve progress grounding through linguistic reasoning~\citep{progressthink} and landmark-centric representations~\citep{dualanchoring}. %However, these methods rely solely on the agent's current and past observations to determine whether a subgoal has been reached or which direction to take. 

To examine whether existing agents can recognize this uncertainty, we compare their decision confidence on failed trajectories and success trajectories (detailed in Appendix~\ref{app:progress_myopia}). If their decision confidence reflects the reliability of progress grounding, it should decrease when the agent begins to deviate from the goal. As shown in Fig.~\ref{fig:first-figure}(c), however, three representative navigation models exhibit similar confidence in both cases. This indicates that they remain confident even after deviating from the goal. We formulate this failure mode as \textbf{\emph{Progress Myopia}}: navigation agents fail to recognize unreliable progress grounding and continue to act based on insufficient evidence.

To address this issue, we propose \textit{\textbf{SeekVLN}}, an evidence-seeking framework that improves progress grounding by combining semantic progress reasoning with visual evidence seeking. Rather than grounding progress based on the historical observations, SeekVLN enables the agent to recognize when the available evidence is insufficient and acquire additional task-relevant observations before executing the next navigation action. In this way, SeekVLN shifts VLN from passive reasoning to active evidence seeking for reliable progress grounding. To learn this behavior, we first introduce \textbf{Future-guided Reverse Generation~(FRG)}, which augments offline expert trajectories with supplementary views and evidence annotations. FRG reasons backward from future expert actions to determine where to seek evidence and annotate what key evidence is. Supervised fine-tuning (SFT) on the resulting trajectories establishes a prior for evidence seeking and progress reasoning without additional expert interaction. However, imitation alone does not reveal whether seeking improves subsequent navigation. We therefore introduce \textbf{Counterfactual Contrastive Policy Optimization~(C2PO)} for reinforcement fine-tuning. When the agent chooses to seek evidence, C2PO additionally rolls out a counterfactual direct-navigation branch from the same state. By comparing the two branches, a contrastive reward assigns credit to the seeking decision based on its subsequent navigation benefit. This signal is combined with an adaptive outcome reward to jointly optimize evidence seeking and navigation.

Our main contributions are summarized as follows:

\begin{itemize}

    \item We identify and formulate \textbf{\emph{Progress Myopia}}, a failure mode in which VLN agents fail to recognize unreliable progress grounding under insufficient evidence. %We examine this behavior through condidence analysis over success and failure trajectories.

    %\item We propose \textbf{SeekVLN}, which combines semantic progress reasoning with active acquisition of task-relevant observations. FRG reasons backward from the future expert's decisions to establish an initial evidence-seeking prior, while C2PO compares evidence-seeking and direct-navigation branches to assign credit to evidence-seeking decisions.
    \item We propose SeekVLN, which combines semantic progress reasoning with active acquisition of task-relevant observations. FRG converts offline demonstrations into evidence-seeking supervision, and supervised fine-tuning establishes an initial evidence-seeking prior. C2PO then applies reinforcement fine-tuning, comparing evidence-seeking and direct-navigation branches from the same state to assign credit to seeking decisions.
    
    %\item SeekVLN improves success rate by \textbf{12.7\%} and \textbf{7.5\%} over base model on R2R-CE and RxR-CE, respectively. Simulated and real-worlds evaluations and qualitative rollouts further demonstrate its ability to actively seek task-relevant evidence for reliable progress grounding.

    \item SeekVLN sets a new state of the art for VLN agents, improving success rate by 12.7\% and 7.5\% over the base model on R2R-CE and RxR-CE, respectively. Simulated and real-world evaluations further demonstrate its human-like ability to actively seek task-relevant evidence for reliable progress grounding.
\end{itemize}

\section{Related Works}
\subsection{VLM-based Vision-Language Navigation}
%[1] X. Xue et al., “OmniNav: A Unified Framework for Prospective Exploration and Visual-Language Navigation,” Jan. 07, 2026, arXiv: arXiv:2509.25687. doi: 10.48550/arXiv.2509.25687.
%[2] J. Zhang et al., “Uni-NaVid: A Video-based Vision-Language-Action Model for Unifying Embodied Navigation Tasks,” Feb. 06, 2025, arXiv: arXiv:2412.06224. doi: 10.48550/arXiv.2412.06224.
%[3] A.-C. Cheng et al., “NaVILA: Legged Robot Vision-Language-Action Model for Navigation,” Feb. 17, 2025, arXiv: arXiv:2412.04453. doi: 10.48550/arXiv.2412.04453.
%[4] M. Wei et al., “StreamVLN: Streaming Vision-and-Language Navigation via SlowFast Context Modeling,” Jul. 07, 2025, arXiv: arXiv:2507.05240. doi: 10.48550/arXiv.2507.05240.
%[5] J. Zhang et al., “Embodied Navigation Foundation Model,” Sep. 16, 2025, arXiv: arXiv:2509.12129. doi: 10.48550/arXiv.2509.12129.
%[6] C. Gao et al., “OctoNav: Towards Generalist Embodied Navigation,” Jun. 11, 2025, arXiv: arXiv:2506.09839. doi: 10.48550/arXiv.2506.09839.
The concept of VLN was introduced as a visually grounded instruction-following problem in real 3D environments~\citep{anderson2018vision}. Before the recent adoption of Vision-Language Models (VLMs), representative VLN agents~\citep{hong2021vln,hamt,duet,an2024etpnav} improved cross-modal state modeling and long-horizon planning through traditional deep learning. Recent VLM-based approaches~\citep{uninavid,navila,streamvln,omninav,octonav,navfom} further broaden the capabilities of navigation agents by leveraging video understanding and action modeling. 
%Uni-NaVid~\cite{uninavid} introduces a unified video-based Vision-Language-Action~(VLA) model for diverse embodied navigation tasks, while NaVILA~\cite{navila} connects high-level VLA outputs with low-level legged locomotion skills for real-world robot navigation. StreamVLN~\cite{streamvln} improves online decision making by maintaining slow-fast context over streaming visual observations and action histories. OmniNav~\cite{omninav}, OctoNav~\cite{octonav}, and NavFoM~\cite{navfom} pursue more general navigation foundations by unifying prospective exploration, multiple goal specifications, diverse navigation tasks, or cross-embodiment data within a single framework.
However, existing VLM policies passively reason and navigate from available observations, creating an information bottleneck when task-relevant evidence is missing. SeekVLN instead actively seeks additional evidence before navigation.
% However, these VLM policies largely ground task progress from the observed trajectory and the current egocentric view. When a landmark is outside the field of view or the instruction \zm{is ambiguous} 
% %under-specifies a fork
% , the agent \zm{may} decide under insufficient evidence. SeekVLN is complementary to this line of work: instead of only strengthening the VLM policy, it explicitly learns when to acquire additional task-relevant observations before \zm{executing a
% navigation action}.
%committing to a navigation action.

\subsection{Progress Reasoning and Grounding in Embodied Navigation}
%[1] S. Wang et al., “Progress-Think: Semantic Progress Reasoning for Vision-Language Navigation,” Nov. 21, 2025, arXiv: arXiv:2511.17097. doi: 10.48550/arXiv.2511.17097.
%[2] K. Wu et al., “Dual-Anchoring: Addressing State Drift in Vision-Language Navigation,” 2026, arXiv. doi: 10.48550/ARXIV.2604.17473.
%[3] W. Guo et al., “AwareVLN: Reasoning with Self-awareness for Vision-Language Navigation,” May 21, 2026, arXiv: arXiv:2605.22816. doi: 10.48550/arXiv.2605.22816.
%[1] H. Wang, W. Wang, T. Shu, W. Liang, and J. Shen, “Active Visual Information Gathering for Vision-Language Navigation,” in Computer Vision – ECCV 2020, vol. 12367, A. Vedaldi, H. Bischof, T. Brox, and J.-M. Frahm, Eds., in Lecture Notes in Computer Science, vol. 12367. , Cham: Springer International Publishing, 2020, pp. 307–322. doi: 10.1007/978-3-030-58542-6_19.
%[1] X. Ding et al., “AdaNav: Adaptive Reasoning with Uncertainty for Vision-Language Navigation,” Sep. 29, 2025, arXiv: arXiv:2509.24387. doi: 10.48550/arXiv.2509.24387.
%ref to more related works of progress-think
Progress grounding has long been a concern in VLN: agents must determine both where to move and which subgoals have been completed. Early agents used progress estimates for visual-textual grounding, backtracking, and action selection~\citep{selfmonitoring,regretful}, and explored surrounding viewpoints to reduce ambiguity before navigation~\citep{activevig}. Recent work revisits this issue with richer reasoning mechanisms. Progress-Think~\citep{progressthink} predicts semantic progress, Dual-Anchoring~\citep{dualanchoring} addresses progress and memory drift, and AdaNav~\citep{adanav} and AwareVLN~\citep{awarevln} introduce uncertainty-aware or self-aware reasoning.
% These approaches make the agent's internal task state more interpretable, but they still largely reason over evidence that has already been observed. As a result, a confident progress estimate can still be unreliable when the missing evidence is visually unavailable. SeekVLN differs by formulating this failure as Progress Myopia and coupling semantic progress reasoning with evidence seeking, so that the agent can actively obtain the observations needed to ground its progress.
These approaches improve progress grounding through interpretable evaluation modules, but overlook the unreliability caused by environmental uncertainty. SeekVLN formulates this limitation as \textbf{\emph{Progress Myopia}} and introduces an evidence-seeking framework for more reliable progress grounding.

\subsection{Reinforcement Learning for Vision-Language Navigation}
%[1] Z. Qi, Z. Zhang, Y. Yu, J. Wang, and H. Zhao, “VLN-R1: Vision-Language Navigation via Reinforcement Fine-Tuning,” Jun. 25, 2025, arXiv: arXiv:2506.17221. doi: 10.48550/arXiv.2506.17221.
%[2] S. Ye et al., “ETP-R1: Evolving Topological Planning with Reinforcement Fine-tuning for Vision-Language Navigation in Continuous Environments,” Dec. 24, 2025, arXiv: arXiv:2512.20940. doi: 10.48550/arXiv.2512.20940.
%[3] Q. Liu, T. Huang, Z. Zhang, and H. Tang, “Nav-R1: Reasoning and Navigation in Embodied Scenes,” Sep. 13, 2025, arXiv: arXiv:2509.10884. doi: 10.48550/arXiv.2509.10884.
%[4] Z. Zhang et al., “ActiveVLN: Towards Active Exploration via Multi-Turn RL in Vision-Language Navigation,” Sep. 16, 2025, arXiv: arXiv:2509.12618. doi: 10.48550/arXiv.2509.12618.
%[5] H. Li, R. Liu, H. Fan, and Y. Yang, “Let’s Reward Step-by-Step: Step-Aware Contrastive Alignment for Vision-Language Navigation in Continuous Environments,” Mar. 10, 2026, arXiv: arXiv:2603.09740. doi: 10.48550/arXiv.2603.09740.
%[6] Z. Wang, Z. Lin, Y. Yang, H. Fu, and D. Ye, “SeeNav-Agent: Enhancing Vision-Language Navigation with Visual Prompt and Step-Level Policy Optimization,” Dec. 02, 2025, arXiv: arXiv:2512.02631. doi: 10.48550/arXiv.2512.02631.
Unlike imitation learning, RL optimizes VLN policies on their own rollouts, exposing recovery and exploration actions to task-level feedback~\citep{activevln,stepaware}. Recent methods apply RL to VLM navigation: VLN-R1~\citep{vlnr1} uses GRPO-style training, Nav-R1~\citep{navr1} emphasizes embodied reasoning, and ETP-R1~\citep{etpr1} and ActiveVLN~\citep{activevln} extend RL to topological planning and active exploration. To address sparse feedback, \citet{stepaware} introduce step-level contrastive rewards, while SeeNav-Agent~\citep{seenav} combines visual prompting with step-level policy optimization.
% Nevertheless, existing objectives mainly evaluate final success, path progress or reasoning quality; they rarely isolate whether gathering extra visual evidence is better than navigating directly from the same state. SeekVLN addresses this credit-assignment gap with Future-guided Reverse Generation and Counterfactual Contrastive Policy Optimization. This contrastive signal explicitly optimizes both when to seek evidence and how to use the acquired evidence for reliable progress grounding.
Evidence seeking poses a credit-assignment challenge because its benefit emerges through later navigation. C2PO addresses this by comparing matched evidence-seeking and direct-navigation branches and rewarding the difference in short-horizon progress.

\section{Methodology}

We propose \textbf{SeekVLN}, an evidence-seeking framework for reliable progress grounding in VLN. As illustrated in Fig.~\ref{fig:framework}, SeekVLN assesses whether the current visual context is sufficient and acquires supplementary views when needed before reasoning about progress and acting. We train this behavior in two stages. First, Future-guided Reverse Generation (FRG) constructs evidence-seeking supervision from offline expert trajectories, establishing a behavior prior. Second, Counterfactual Contrastive Policy Optimization (C2PO) jointly optimizes evidence seeking and navigation through reinforcement fine-tuning.

\subsection{Evidence-Seeking Navigation Framework}

\paragraph{Problem formulation.}
We study monocular Vision-and-Language Navigation in Continuous Environments (VLN-CE)~\citep{DBLP:conf/eccv/KrantzWMBL20}, where an agent navigates through a continuous 3D environment according to a natural-language instruction $I$. At decision step $t$, the agent receives the current observation $o_t$ and a sampled visual history $\mathcal{H}_t=\{o_{i_m}\}_{m=1}^{M}$ comprising $M$ historical frames. The policy directly conditions on these inputs to generate the decision sequence
\begin{equation}
    y_t = \pi_\theta\!\left(I, o_t, \mathcal{H}_t\right).
\end{equation}

\begin{figure}[t]
    \centering
    \includegraphics[width=\columnwidth]{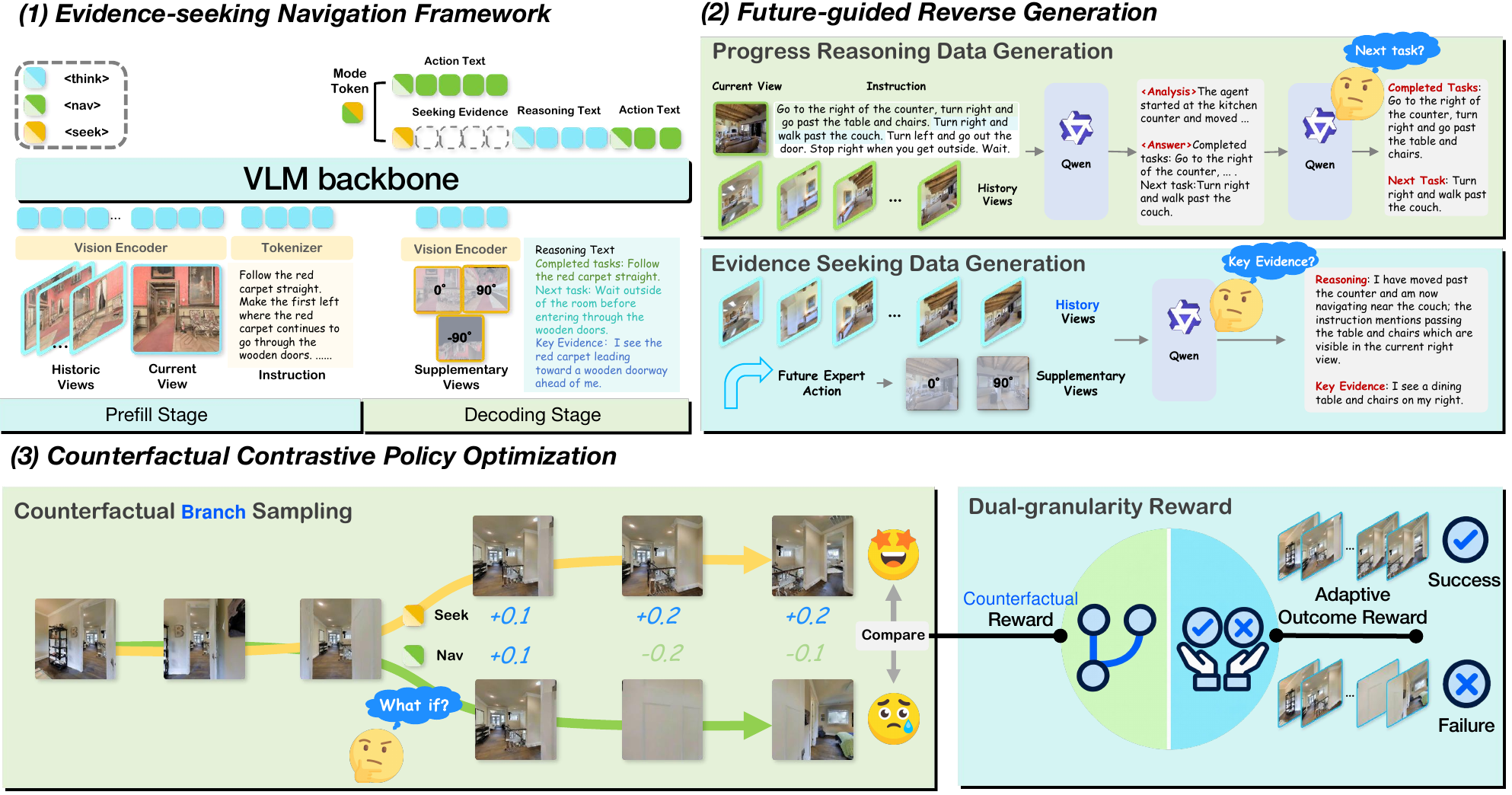}
    \caption{Overview of \textbf{SeekVLN}. (1) The policy decides whether to navigate or seek evidence. (2) FRG reasons backward from future expert actions to determine where to seek evidence and annotate key evidence, while using historical context for progress annotation. (3) C2PO compares evidence-seeking and direct-navigation branches and jointly optimizes evidence-seeking and navigation.}
    \label{fig:framework}
    \vspace{-0.2cm}
\end{figure}

\paragraph{Dual-mode navigation.}
A conventional VLN policy directly predicts a navigation action at every state. SeekVLN instead begins each decision by determining whether the available observations provide sufficient evidence for reliable progress grounding. As illustrated in Fig.~\ref{fig:framework}(1), the first control token selects one of two modes:
\begin{equation}
m_t =
\begin{cases}
    \mathrm{SEEK}, & y_t^{(1)}=\texttt{<seek>},\\
    \mathrm{NAV},  & y_t^{(1)}=\texttt{<nav>}.
\end{cases}
\end{equation}

Here, $y_t^{(1)}$ denotes the first token of the response. If the agent selects \textsc{NAV}, it directly predicts the navigation action, $y_t=\texttt{<nav>}\oplus y_t^{\mathrm{nav}}$, where $\oplus$ denotes token concatenation. If the agent selects \textsc{SEEK}, it first acquires supplementary views and reasons over them to seek evidence and estimate progress before predicting the navigation action.

\paragraph{Evidence-seeking interaction.}
After the policy predicts \texttt{<seek>}, the environment appends left, front, and right views at fixed relative headings, forming the interaction prefix
\begin{equation}
    y_t^{\mathrm{seek}}
    =
    \texttt{<seek>}
    \oplus
    \left[
    o_t^{-90^\circ},
    o_t^{0^\circ},
    o_t^{90^\circ}
    \right]
    \oplus
    \texttt{</seek>},
\end{equation}
Conditioned on the supplementary views, the policy generates \texttt{<think>} to begin the progress-reasoning segment
\begin{equation}
    y_t^{\mathrm{prog}}
    =
    \texttt{<think>}
    \oplus
    \left(
    g_t^{-},
    g_t^{+},
    e_t
    \right)
    \oplus
    \texttt{</think>},
\end{equation}
where $g_t^{-}$ summarizes completed instruction subgoals, $g_t^{+}$ specifies the next subgoal, and $e_t$ identifies the visual evidence supporting progress grounding. Fig.~\ref{fig:framework}(1) shows a concrete example of these three parts. The complete \textsc{SEEK}-mode sequence is
\begin{equation}
    y_t
    =
    y_t^{\mathrm{seek}}
    \oplus
    y_t^{\mathrm{prog}}
    \oplus
    \texttt{<nav>}\oplus
    y_t^{\mathrm{nav}}.
\end{equation}

\subsection{Future-Guided Reverse Generation}
Learning evidence seeking requires supervision for when to acquire additional observations and what task-relevant evidence they provide. However, such annotations are not explicitly available in offline expert trajectories. We therefore introduce \textbf{Future-guided Reverse Generation (FRG)}, which reasons backward from future expert actions to construct evidence-seeking supervision without additional expert interaction. Specifically, FRG addresses two questions: \emph{(1) When and where should the agent seek evidence? (2) What is the key evidence in the selected views?} 

\paragraph{Evidence-seeking targets.}
Future expert actions provide a cue for when and where additional evidence may be useful. For example, a side view may help ground an upcoming turn, whereas the front view is enough for moving along a straight hallway. Based on this observation, FRG assigns action-dependent target proportions and deterministically selects decisions for \textsc{SEEK} supervision, producing the mode plan $\mathbf{m}^{*}$ (detailed in Appendix~\ref{app:hrg}).

For each selected \textsc{SEEK} state, FRG uses the upcoming expert action to determine where to look for evidence. The annotator VLM then receives the corresponding subset of current views, together with the instruction and observed history, and identifies the task-relevant evidence:
\begin{equation}
    e_t = f^{\mathrm{ev}}_{\mathrm{VLM}}\!\left(I,\mathcal{H}_t^{*},\mathcal{V}_t(a_t^{*})\right),
\end{equation}
where $\mathcal{V}_t(a_t^{*})$ denotes the current views selected according to the expert action $a_t^{*}$.

\paragraph{Progress-reasoning targets.}
As illustrated in Fig.~\ref{fig:framework} (2), given the instruction, the historical
and current observations, and the provided subgoal list $\mathcal{S}(I)$, the annotator VLM first analyzes the current task progress, then distills the analysis into structured progress fields:
\begin{equation}
    \left(g_t^{-},\, g_t^{+}\right) = f^{\mathrm{prog}}_{\mathrm{VLM}}\!\left(I,\mathcal{H}_t^{*},o_t^{*},\mathcal{S}(I)\right),
\end{equation}
Together with $e_t$, the annotation is serialized into
$y_t^{\mathrm{prog}}=\texttt{<think>}\oplus(g_t^{-},g_t^{+},e_t)\oplus\texttt{</think>}$.

\paragraph{Prior dataset construction.}
Given the input context
$x_t^{*}=(I, o_t^{*}, \mathcal{H}_t^{*})$, a \textsc{NAV} state is paired
with the expert navigation action $y_t^{\mathrm{nav},*}$, and a \textsc{SEEK} state
with the full response of evidence seeking, progress reasoning, and navigation action. The prior
dataset is
\begin{equation}
\begin{aligned}
\mathcal{D}_{\mathrm{prior}}
&= \mathcal{D}_{\mathrm{nav}} \cup \mathcal{D}_{\mathrm{seek}},\\
\label{eq:prior_dataset}
\mathcal{D}_{\mathrm{nav}}
&= \{(x_t^{*},\texttt{<nav>}\oplus y_t^{\mathrm{nav},*})\mid m_t^{*}=\mathrm{NAV}\},\\
\mathcal{D}_{\mathrm{seek}}
&= \{(x_t^{*},y_t^{\mathrm{seek}}\oplus y_t^{\mathrm{prog}}\oplus \texttt{<nav>}\oplus y_t^{\mathrm{nav},*})
\mid m_t^{*}=\mathrm{SEEK}\}.
\end{aligned}
\end{equation}
$\mathcal{D}_{\mathrm{prior}}$ thus jointly supervises direct navigation,
evidence seeking, and progress reasoning.

\subsection{Counterfactual Contrastive Policy Optimization}
\label{method:c2po}

Supervised fine-tuning on $\mathcal{D}_{\mathrm{prior}}$ initializes a basic behavior prior for triggering evidence seeking and reasoning over acquired views. However, imitation alone does not reveal whether a seeking decision improves subsequent navigation under on-policy interaction.

We therefore introduce \textbf{Counterfactual Contrastive Policy Optimization} (C2PO) to address two coupled questions: \emph{(1) How can a policy recognize insufficient evidence and trigger evidence seeking adaptively? (2) How can evidence seeking and navigation be optimized jointly?} C2PO compares evidence-seeking and direct-navigation branches from the same state to assign credit to a seeking decision. The resulting contrastive reward is combined with an adaptive outcome reward, providing dual-granularity feedback to jointly optimize evidence seeking and navigation.

\paragraph{Counterfactual branch sampling.}
A sampled trajectory reveals the outcome of the chosen mode, but not what would have happened under the alternative. To assess the benefit of a seeking decision, we therefore compare evidence seeking with direct navigation from the same state. As illustrated in
Fig.~\ref{fig:framework} (3), when the main rollout selects \textsc{SEEK} under context $(I,o_t,\mathcal{H}_t)$, we clone the simulator state, observation history, and model context, then compare the factual evidence-seeking branch
\begin{equation}
    \tau_{t:t+H}^{\mathrm{seek}}
    \sim
    \pi_\theta\!\left(
        \cdot \mid I,o_t,\mathcal{H}_t,
        m_t=\mathrm{SEEK}
    \right).
\end{equation}
The counterfactual branch starts from the same state, with the first mode decision forced to \textsc{NAV}:
\begin{equation}
    \tau_{t:t+H}^{\mathrm{nav}}
    \sim
    \pi_\theta\!\left(
        \cdot \mid I,o_t,\mathcal{H}_t,
        m_t=\mathrm{NAV}
    \right).
\end{equation}
Both branches are rolled out with the same policy for $H$ primitive actions, isolating the local benefit of evidence seeking while controlling for the initial state and context.

\paragraph{Dual-granularity reward.}
We combine a decision-level counterfactual reward with an episode-level adaptive outcome reward to jointly optimize evidence seeking and navigation. At decision $t$, the total reward is
\begin{equation}
    r_t = r_t^{\mathrm{cf}} + r_t^{\mathrm{out}}.
\end{equation}

The \emph{counterfactual reward} measures the short-horizon navigation benefit of an evidence-seeking decision. Let $\Delta d_h^b$ denote the normalized reduction in geodesic distance to the goal at primitive action $h$ in branch $b \in \{\mathrm{seek},\mathrm{nav}\}$. We compute the discounted progress difference between the two branches and define
\begin{equation}
    r_t^{\mathrm{cf}} =
    \begin{cases}
        w\,\operatorname{clip}\!\left(
            \displaystyle\sum_{h=1}^{H}
            \gamma_{\mathrm{cf}}^{h-1}
            \left(
                \Delta d_h^{\mathrm{seek}}
                - \Delta d_h^{\mathrm{nav}}
            \right),
            -0.2,\,0.2
        \right),
        & m_t = \mathrm{SEEK}, \\[6pt]
        0,
        & m_t = \mathrm{NAV},
    \end{cases}
\end{equation}
where $\gamma_{\mathrm{cf}}$ discounts later progress and $w$ controls the contribution of the counterfactual reward. 
%Let $\Delta d_h^b$ denote normalized geodesic progress at primitive action $h$ in branch $b\in\{\mathrm{NAV},\mathrm{SEEK}\}$. We define
% \begin{equation}
% r_t^{\mathrm{cf}}=
% \begin{cases}
% w\,\operatorname{clip}\!\left(
% \displaystyle\sum_{h=1}^{H}\gamma_{\mathrm{cf}}^{h-1}
% \bigl(\Delta d_h^{\mathrm{seek}}-\Delta d_h^{\mathrm{nav}}\bigr),
% -0.2,0.2\right), & m_t=\mathrm{SEEK},\\
% 0, & m_t=\mathrm{NAV},
% \end{cases}
% \end{equation}
% The signal is zero for direct-navigation decisions; $w$ controls its contribution.

The \emph{adaptive outcome reward} is given once at episode termination:
\begin{equation}
r_t^{\mathrm{out}}
= \mathbb{I}[t=T]
\begin{cases}
1+0.2\,\mathrm{SPL}, & \text{if success},\\
-0.5, & \text{otherwise},
\end{cases}
\end{equation}
where $T$ denotes the terminal decision.

%We estimate advantages and optimize the policy and value function with PPO~\citep{schulman2017proximalpolicyoptimizationalgorithms}. Counterfactual branches are used only during reinforcement fine-tuning and incur no evaluation-time rollout cost.
We use the combined rewards to estimate advantages and optimize the policy and value function with PPO~\citep{schulman2017proximalpolicyoptimizationalgorithms}. Counterfactual branches are sampled only during reinforcement fine-tuning and require no additional branch rollouts during evaluation.

\suppressfloats[t]
\section{Experiment}

\begin{table*}[t]
\centering
\caption{
Comparison of different methods on the R2R-CE and RxR-CE Val-Unseen splits.
Observations include a single RGB camera (S.RGB) and depth sensor (Depth).
$\dagger$ indicates methods without using LLMs. %\codexedit{CM$^2$ results on RxR-CE are from \citet{wang2024sim}.}
}
\label{tab:r2r_rxr_ce_val_unseen}

\small
\setlength{\tabcolsep}{3pt}
\renewcommand{\arraystretch}{1.02}

\resizebox{\textwidth}{!}{%
\begin{tabular}{@{}lcc|cccc|cccc@{}}
\toprule
\multirow{2}{*}{Method}
& \multicolumn{2}{c|}{Observation}
& \multicolumn{4}{c|}{R2R-CE Val-Unseen}
& \multicolumn{4}{c}{RxR-CE Val-Unseen} \\
\cmidrule(lr){2-3}
\cmidrule(lr){4-7}
\cmidrule(l){8-11}
& S.RGB
& Depth
& NE$\downarrow$
& OSR$\uparrow$
& SR$\uparrow$
& SPL$\uparrow$
& NE$\downarrow$
& SR$\uparrow$
& SPL$\uparrow$
& nDTW$\uparrow$ \\
\midrule

BEVBert$^{\dagger}$~\cite{an2022bevbert}
&  & \checkmark
& 4.57 & 67.0 & 59.0 & 50.0
& -- & -- & -- & -- \\

ETPNav$^{\dagger}$~\cite{an2024etpnav}
&  & \checkmark
& 4.71 & 65.0 & 57.0 & 49.0
& 5.64 & 54.8 & 44.9 & 61.9 \\

ENP-ETPNav$^{\dagger}$~\cite{liu2024vision}
&  & \checkmark
& 4.69 & 65.0 & 58.0 & 50.0
& 5.51 & 55.3 & 45.1 & 63.0 \\

\midrule

Seq2Seq$^{\dagger}$~\cite{DBLP:conf/eccv/KrantzWMBL20}
& \checkmark & \checkmark
& 7.77 & 37.0 & 25.0 & 22.0
& 12.10 & 13.9 & 11.9 & -- \\

CMA$^{\dagger}$~\cite{DBLP:conf/eccv/KrantzWMBL20}
& \checkmark & \checkmark
& 7.37 & 40.0 & 32.0 & 30.0
& -- & -- & -- & --  \\

LAW$^{\dagger}$~\cite{raychaudhuri2021language}
& \checkmark & \checkmark
& 6.83 & 44.0 & 35.0 & 31.0
& 10.90 & 8.0 & 8.0 & -- \\

CM$^2$$^{\dagger}$~\cite{georgakis2022cross}
& \checkmark & \checkmark
& 7.02 & 41.0 & 34.0 & 27.0
& 8.98 & 14.4 & 9.2 & -- \\

WS-MGMap$^{\dagger}$~\cite{chen2022weakly}
& \checkmark & \checkmark
& 6.28 & 47.0 & 38.0 & 34.0
& -- & -- & -- & -- \\

sim2real$^{\dagger}$~\cite{wang2024sim}
& \checkmark & \checkmark
& 5.95 & 55.8 & 44.9 & 30.4
& -- & -- & -- & -- \\

NavMorph$^{\dagger}$~\cite{yao2025navmorph}
& \checkmark & \checkmark
& 5.75 & 56.9 & 47.9 & 33.2
& 8.85 & 30.8 & 22.8 & 44.2 \\

NaVid-4D~\cite{liu2025vid}
& \checkmark & \checkmark
& 5.99 & 55.7 & 43.8 & 37.1
& -- & -- & -- & -- \\

\midrule

NaVid~\cite{zhang2024navid}
& \checkmark &
& 5.47 & 49.1 & 37.4 & 35.9
& -- & -- & -- & -- \\

Uni-NaVid~\cite{uninavid}
& \checkmark &
& 5.58 & 53.5 & 47.0 & 42.7
& 6.24 & 48.7 & 40.9 & -- \\

NaVILA~\cite{navila}
& \checkmark &
& 5.22 & 62.5 & 54.0 & 49.0
& 6.77 & 49.3 & 44.0 & 58.8 \\

NavFoM~\cite{navfom}
& \checkmark &
& 5.01 & 64.9 & 56.2 & 51.2
& \underline{5.51} & \underline{57.4} & \underline{49.4} & 60.2 \\

StreamVLN~\cite{streamvln}
& \checkmark &
& 4.98 & 64.2 & 56.9 & 51.9
& 6.22 & 52.9 & 46.0 & 61.9 \\

Progress-Think~\cite{progressthink}
& \checkmark &
& \underline{4.68} & 63.6 & 60.1 & 53.6
& -- & -- & -- & -- \\

\rowcolor{blue!3}
Aux-Think~\cite{auxthink} (base model)
& \checkmark &
& 6.08 & 60.0 & 54.8 & 46.9
& 6.24 & 52.2 & 40.2 & -- \\

\rowcolor{blue!8}
SeekVLN-FRG-SFT (ours)
& \checkmark &
& 4.7 & \underline{68.4} & \underline{61.0} & \underline{55.9}
& 5.8 & 55.7 & 47.4 & \underline{62.3} \\

\rowcolor{blue!15}
\textbf{SeekVLN-C2PO-RFT (ours)}
& \checkmark &
& \textbf{3.7} & \textbf{75.2} & \textbf{67.5} & \textbf{61.4}
& \textbf{4.9} & \textbf{59.7} & \textbf{50.3} & \textbf{63.6} \\
\bottomrule
\end{tabular}}
\end{table*}

\subsection{Experimental Setting}
\label{sec:experimental-settings}

\paragraph{Simulation Environments and Metrics.}
We conduct experiments in the Habitat simulator~\citep{DBLP:conf/iccv/SavvaMPBKMZWJSL19} using Matterport3D scenes~\citep{DBLP:conf/3dim/ChangDFHNSSZZ17}, and evaluate on the val-unseen splits of R2R-CE~\citep{DBLP:conf/eccv/KrantzWMBL20} and RxR-CE~\citep{DBLP:conf/emnlp/KuAPIB20}. Following their standard protocols, we report Navigation Error (NE), Oracle Success Rate (OSR), Success Rate (SR), and Success weighted by Path Length (SPL) on R2R-CE, and NE, SR, SPL, and normalized Dynamic Time Warping (nDTW) on RxR-CE. NE measures the final geodesic distance to the goal; OSR measures the fraction of trajectories that enter the success radius; SR evaluates success at the final position; SPL measures success weighted by navigation efficiency; and nDTW measures fidelity to the reference trajectory. %Ablations report NE, SR, and SPL unless otherwise specified.

\paragraph{Model training.}
We initialize SeekVLN from the pretrained Aux-Think model~\citep{auxthink} and train it in two stages: SFT on the FRG prior dataset, constructed from 4K R2R-CE train split episodes with Qwen-VL-Max~\citep{qwen-vl}, followed by C2PO reinforcement fine-tuning on 640 sampled R2R-CE training episodes. Details of both stages are provided in Appendix~\ref{app:training_details}.

\paragraph{Action space.} We use the same action space as NaVILA~\citep{navila}. It formulates continuous navigation using a discrete high-level action space consisting of \{\texttt{move forward}, \texttt{turn left}, \texttt{turn right}, \texttt{stop}\}. Forward actions move the agent by 25, 50, or 75 centimeters, whereas left and right turns rotate it by $15^{\circ}$, $30^{\circ}$, or $45^{\circ}$.

\subsection{Main Results}
\label{sec:main_results}

Table~\ref{tab:r2r_rxr_ce_val_unseen} compares SeekVLN with representative VLN methods on the R2R-CE and RxR-CE Val-Unseen splits. Both training stages contribute: FRG-SFT already surpasses the base model and most prior methods, and C2PO-RFT further sets a new state of the art on both benchmarks.

\paragraph{FRG-SFT results.}
Fine-tuning on the FRG prior dataset alone brings clear gains over the base model. On R2R-CE, SeekVLN-FRG-SFT raises SR from 54.8 to 61.0 and SPL from 46.9 to 55.9; on RxR-CE, SR improves from 52.2 to 55.7 and SPL from 40.2 to 47.4. These gains come from offline supervision only, requiring no additional environment interaction or expert queries. Notably, SPL gains exceed SR gains on both benchmarks, suggesting that evidence seeking primarily improves path efficiency by reducing unnecessary actions caused by unreliable decisions. With SFT alone, SeekVLN already outperforms prior VLM agents such as NaVILA, StreamVLN, and NavFoM on R2R-CE, and exceeds Progress-Think, which likewise adds explicit progress reasoning.

\paragraph{Gains from C2PO-RFT.}
Reinforcement fine-tuning with C2PO improves all metrics on both benchmarks.
On R2R-CE, SR increases from 61.0 to 67.5 ($+6.5$) and SPL from 55.9 to 61.4
($+5.5$) over the SFT checkpoint, achieving the best among all compared
methods. On RxR-CE, SR reaches 59.7 ($+4.0$) and SPL 50.3 ($+2.9$),
surpassing NavFoM~\citep{navfom}, the strongest prior VLM agent, by 2.3 and 0.9 points.
Since SFT has already established basic evidence-seeking behavior, these
gains indicate that C2PO mainly improves \emph{when} to seek and \emph{how}
to use the acquired evidence. The counterfactual reward plays a key role: it
credits a seeking decision only when the acquired evidence improves
subsequent navigation, which discourages unnecessary triggers and reinforces
useful ones. RxR-CE benefits particularly from this credit assignment, as
its longer and more fine-grained instructions require reliable progress
grounding over longer horizons. 

Overall, compared with the base model, the full two-stage training yields total gains of $+12.7$ SR and $+14.5$ SPL on R2R-CE, and $+7.5$ SR and $+10.1$ SPL on RxR-CE, establishing a new state of the art among VLM-based agents on both benchmarks.

\paragraph{Ablation studies.}
%Appendix~\ref{sec:ablation} isolates the contributions of FRG supervision and the C2PO counterfactual reward.
Appendix~\ref{sec:ablation} presents controlled ablation studies to isolate the respective contributions of FRG supervision and the C2PO counterfactual reward to navigation performance.

\subsection{Deep Dive into Evidence-Seeking Behavior}
\label{sec:evidence_seeking_analysis}

Beyond aggregate navigation metrics, we organize the behavioral analysis around three questions: \emph{(1) Does adaptive triggering outperform fixed triggering strategies? (2) Does the acquired evidence improve subsequent navigation decisions? (3) Does SeekVLN exhibit human-like evidence-seeking behavior?} We answer these questions through inference-time interventions, direct behavioral metrics, and qualitative rollout visualizations, respectively.

\begin{figure}[H]
    \centering
    \includegraphics[width=0.98\textwidth]{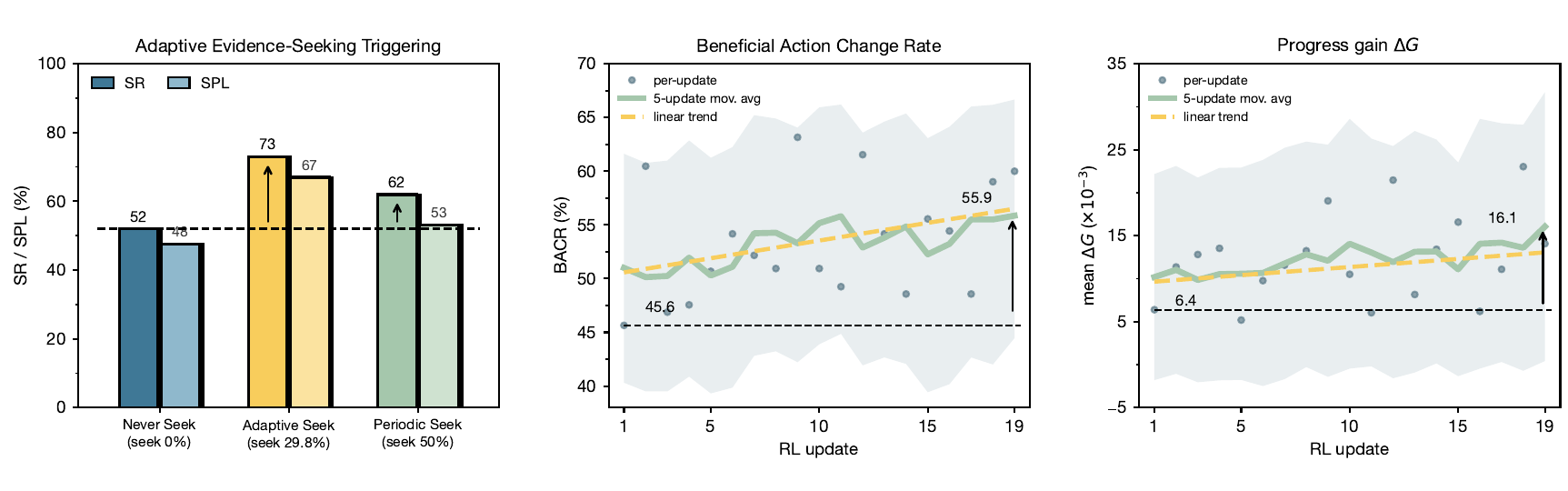}
    \caption{\textbf{Deep analysis of evidence-seeking behavior.}
Left: navigation success and path efficiency under three trigger strategies, highlighting the effectiveness of adaptive evidence seeking.
Center: beneficial action change rate (BACR) during reinforcement fine-tuning, measuring the proportion of beneficial action changes. Right: mean short-horizon progress gain of the evidence-seeking branch over the direct-navigation branch, quantifying the navigation benefit of acquired evidence.}
    \label{fig:deep_dive}
\end{figure}

\paragraph{Adaptive Evidence-Seeking Triggering.}
\label{sec:adaptive_triggering}

All triggering experiments are conducted on a 100-episode subset sampled
from the R2R-CE Val-Unseen split, covering diverse scenes and trajectories
with varied task complexity. The strategies are realized by manually
prefilling the mode token of the same trained policy (SeekVLN-C2PO-RFT):
\emph{Never Seek} forces \texttt{<nav>} at every decision, \emph{Periodic
Seek} prefills \texttt{<seek>} every two steps with \texttt{<nav>} at all
remaining decisions, and \emph{Adaptive Seek} lets the policy make its own
mode decision. 

We compare their seek rates and navigation performance in
Fig.~\ref{fig:deep_dive} (left): Adaptive Seek (29.8\% seek rate) achieves
73\% SR and 67\% SPL, versus 52\%/48\% for Never Seek (0\% seek rate) and 62\%/53\% for
Periodic Seek (50\% seek rate). These results indicate that C2PO-trained SeekVLN
learns when evidence seeking is needed, achieving the strongest navigation performance among the tested triggering strategies without
relying on more frequent seeking.

\paragraph{Effectiveness of Evidence Seeking.}
\label{sec:evidence_usefulness}
Evidence seeking is useful only if the subsequent navigation decisions are improved, rather than merely changing the agent's response. We therefore define the Beneficial Action Change Rate (BACR) as
\begin{equation}
    \mathrm{BACR}
    = \frac{N_{\mathrm{beneficial}}}{N_{\mathrm{seek}}},
\end{equation}
where $N_{\mathrm{seek}}$ is the number of triggered states, and $N_{\mathrm{beneficial}}$ counts those at which seeking changes the next action and yields greater short-horizon progress than direct navigation from the same state. Figure~\ref{fig:deep_dive} (center) shows BACR increasing from 45.6\% to 55.9\% over the RL updates. To measure the magnitude of this benefit, we define the progress gain at a triggered state as
\begin{equation}
    \Delta G_t = \sum_{h=1}^{H}
    \left(\Delta d_h^{\mathrm{seek}}-\Delta d_h^{\mathrm{nav}}\right),
\end{equation}
where $\Delta d_h^b$ denotes normalized geodesic progress at primitive action $h$ in branch $b$, as defined in Section~\ref{method:c2po}. The mean $\Delta G_t$ across triggered states rises from $6.4\times10^{-3}$ to $16.1\times10^{-3}$ (right). Together, the two trends suggest that C2PO learns not only to make more beneficial action changes after seeking, but also to increase their subsequent navigation advantage.%As shown in Fig.~\ref{fig:deep_dive} (center), BACR increases from 45.6\% to 55.9\% over the plotted RL updates. The mean progress advantage of the seek branch likewise rises from $6.4\times10^{-3}$ to $16.1\times10^{-3}$ (right). These trends suggest that C2PO increasingly favors evidence-seeking decisions that lead to more effective navigation, not merely different actions.

\begin{figure}[H]
    \centering
    \includegraphics[width=0.88\columnwidth,trim=0 0pt 0 0pt,clip]{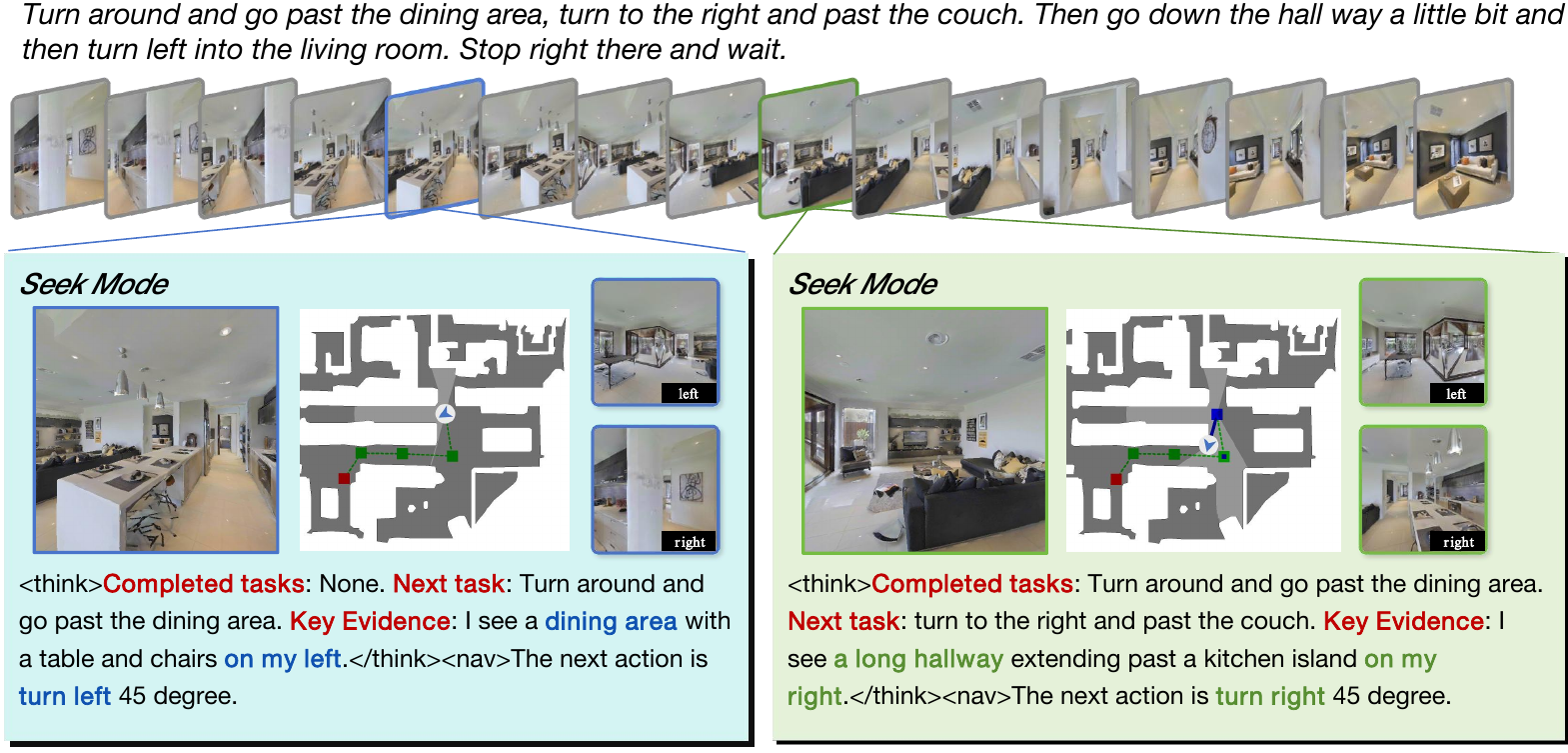}
    \caption{\textbf{Evidence seeking in simulated navigation.} Two SEEK decisions along a representative trajectory. SeekVLN first identifies the dining area on the left; after completing that subgoal, it locates the hallway beyond the kitchen island on the right. In both cases, evidence seeking resolves uncertainty about instruction progress and grounds the next navigation action.}
    \label{fig:evidence_seeking_cases}
\end{figure}

\begin{figure}[H]
    \centering
    \includegraphics[width=0.98\columnwidth,trim=0 0pt 0 0,clip]{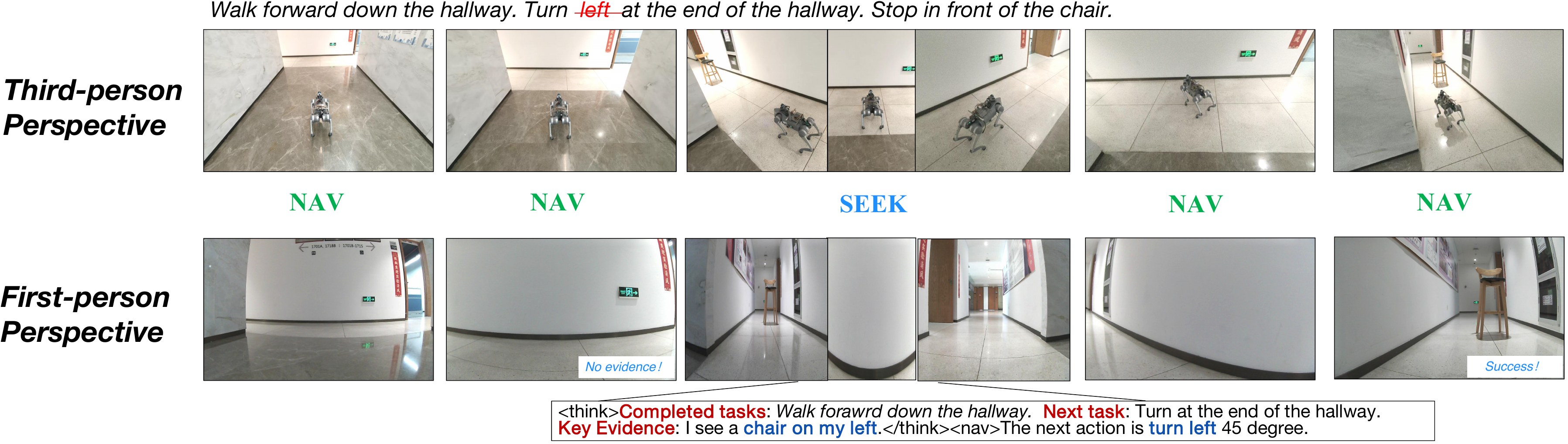}
    \caption{\textbf{Evidence seeking in real-world navigation.} At the end of the hallway, the destination is not visible ahead and the instruction leaves the turn direction unspecified. SeekVLN seeks task-relevant evidence, identifies the chair to its left, and completes the task.}
    \label{fig:real_world_rollout}
\end{figure}

% Temporarily omitted: this analysis is not one of the three questions posed at
% the beginning of the Deep Dive section.
% \subsubsection{Adaptive Trigger Dynamics}
% \label{sec:adaptive_trigger}
%
% We further investigate how the evidence-seeking policy evolves during reinforcement learning.
%
% \paragraph{\textit{Trigger dynamics during training.}}
% Figure~\ref{fig:trigger_analysis}(a) reports the proportion of navigation states at which evidence seeking is triggered over C2PO updates. The trigger rate changes rapidly during the early stage of training and gradually stabilizes at approximately \textbf{29\%} on R2R-CE and \textbf{48.9\%} on RxR-CE. Meanwhile, navigation performance continues to improve and converges after approximately \textbf{XX} updates.
%
% Importantly, C2PO does not simply maximize the frequency of evidence seeking. The stabilized trigger rate therefore reflects a balance between acquiring sufficient task-relevant evidence and maintaining efficient navigation decisions. Together with the improvements in SR and SPL, this trend suggests that the policy learns to trigger evidence seeking more selectively rather than merely increasing its frequency.

\paragraph{Human-Like Evidence-Seeking Behavior.}
\label{sec:evidence_seeking_case}

Figure~\ref{fig:evidence_seeking_cases} shows two targeted evidence-seeking decisions. When the dining area is outside the current view, SeekVLN finds its table and chairs to the left and turns toward it. After completing that subgoal, it locates the hallway beyond the kitchen island on the right and turns accordingly. In both cases, supplementary views resolve a specific uncertainty in progress grounding before action.

Figure~\ref{fig:real_world_rollout} illustrates the same behavior in real-world navigation. At the end of the hallway, the forward view does not reveal the chair, and the instruction does not explicitly specify the left turn. SeekVLN acquires a side view, grounds the destination using the chair on its left, and completes the task. This qualitative success shows that evidence seeking can support navigation under partial observation and instruction ambiguity. The deployment protocol is detailed in Appendix~\ref{app:real-world-deployment}.

\section{Conclusion}
\label{sec:conclusion}

We introduced SeekVLN to address \emph{Progress Myopia} through active evidence seeking for reliable progress grounding. FRG derives a cold-start prior from offline demonstrations, and C2PO learns evidence-seeking decisions through counterfactual credit assignment. Experiments on R2R-CE and RxR-CE demonstrate improved navigation, while simulated and real-world rollouts show targeted evidence-seeking behavior. These results highlight the value of learning when to acquire evidence and how to use it for reliable progress grounding and navigation under partial observation.

\bibliography{seekvln}
\bibliographystyle{iclr2027_conference}

\appendix
\appendixcontents
\section{Detailed Analysis of Progress Myopia}
\appendixsectionentry{Detailed Analysis of Progress Myopia}
\label{app:progress_myopia}

This section provides the analysis underlying Fig.~\ref{fig:first-figure} (c). The analysis asks whether a navigation model's own decision confidence can distinguish reliable decisions on successful trajectories from unreliable decisions on failed trajectories at comparable critical segments.

% \begin{table}[t]
% \centering
% \small
% \caption{Eavaluation protocol for the Progress Myopia analysis. All matching and aggregation are performed separately for each navigation model.}
% \label{tab:progress_myopia_protocol}
% \begin{tabular}{p{0.29\linewidth}p{0.63\linewidth}}
% \toprule
% Item & Protocol \\
% \midrule
% Rollouts & Successful and failed rollouts from the same evaluation split. \\
% Matching & Same instruction when available; otherwise, the same route-length stratum, instruction stage, and critical-event type. \\
% Critical events & Direction-changing decisions and room transitions. \\
% Candidate actions & \texttt{stop}; forward by 25/50/75 cm; or left/right by 15/30/45 degrees. \\
% Reported statistics & Matched-window mean confidence and normalized entropy, together with the number of matched windows and bootstrap confidence intervals. \\
% \bottomrule
% \end{tabular}
% \end{table}

\subsection{Selected Representative Navigation Agents}
\appendixsubsectionentry{Selected Representative Navigation Agents}
We analyze three representative VLM-based navigation agents with different decision mechanisms. \textbf{NaVILA}~\citep{navila} directly predicts high-level navigation actions; \textbf{StreamVLN}~\citep{streamvln} performs streaming action prediction from observation and action histories; and \textbf{Aux-Think}~\citep{auxthink} augments direct navigation with explicit reasoning supervision. Together, these models cover direct action prediction, history-aware navigation, and reasoning-enhanced navigation.

\subsection{Success and Failure Segments Extraction}
\appendixsubsectionentry{Success and Failure Segments Extraction}
We analyze 100 failed rollouts and use their corresponding expert trajectories to construct matched successful counterparts. For each failed rollout, we locate the first segment whose path similarity to the expert route falls below a predefined threshold, focusing the analysis on the onset of deviation rather than its downstream consequences. We align this segment with an equal-length segment at the corresponding position on the expert route. Decision confidence on the latter is measured under teacher forcing, allowing us to compare the model's confidence along its failed continuation with confidence at the matched expert-guided states.

% \subsection{Critical segments}
% \appendixsubsectionentry{Critical segments}
% A critical segment is a fixed decision window centered on a state requiring consequential progress grounding. We consider two recurring event types: (i) \emph{direction-changing decisions}, where the agent must select among competing directions, and (ii) \emph{room transitions}, where the visual context and the active instruction subgoal may change. For a failed rollout, the window is anchored at the first decision after which the trajectory persistently departs from the reference route within that window. Its matched successful window is selected at the corresponding instruction stage and with the same event type. All models use the same window definition and matching protocol.

\subsection{Decision Confidence.}
\appendixsubsectionentry{Decision Confidence}
To compare models with different textual output formats, we score each candidate high-level action $a\in\mathcal{A}$ by its length-normalized sequence log-likelihood,
\begin{equation}
    \ell_t(a)=\frac{1}{|a|}\sum_{i=1}^{|a|}
    \log p\!\left(a_i\mid s_t,a_{<i}\right),
\end{equation}
and normalize these scores over the shared action set $\mathcal{A}$:
\begin{equation}
    p_t(a)=\frac{\exp(\ell_t(a))}{\sum_{a'\in\mathcal{A}}\exp(\ell_t(a'))}.
\end{equation}
Decision confidence is the top-two margin. Let $a_{(1)}$ and $a_{(2)}$ be the actions with the highest and second-highest normalized probabilities, respectively:
\begin{equation}
    c_t=p_t(a_{(1)})-p_t(a_{(2)}).
\end{equation}
We additionally report normalized action entropy as a complementary uncertainty measure. We aggregate both quantities over matched critical windows separately for successful and failed rollouts, and compute confidence intervals by resampling matched windows. Thus, similar confidence and entropy across the two outcomes indicate that the model does not reliably recognize when its progress grounding has become unreliable.

\section{Future-Guided Reverse Generation}
\appendixsectionentry{Future-Guided Reverse Generation}
\label{app:hrg}

% \begin{table}[htbp]
% \centering
% \small
% \caption{FRG data-generation pipeline. Hindsight is an annotation-only signal and is never serialized into a policy input or target.}
% \label{tab:hrg_pipeline}
% \begin{tabular}{p{0.20\linewidth}p{0.36\linewidth}p{0.34\linewidth}}
% \toprule
% Stage & Input & Output / check \\
% \midrule
% Candidate selection & Offline expert trajectory & Decision states covering regular intervals, turns, stage transitions, and terminal states. \\
% Mode annotation & Policy-visible context, three-view bundle, and future expert continuation & \textsc{NAV}/\textsc{SEEK}; SEEK requires missing but useful evidence in the three-view bundle. \\
% Progress annotation & SEEK state, three-view bundle, and hindsight & Completed tasks, next task, and key evidence grounded in the current bundle. \\
% Validation and serialization & Structured annotation and expert macro action & Keep valid labels; serialize a NAV or two-turn SEEK training example. \\
% \bottomrule
% \end{tabular}
% \end{table}

\subsection{Candidate-State Construction}
\appendixsubsectionentry{Candidate-State Construction}
\label{app:hrg:candidates}
FRG retains every macro decision in an offline expert trajectory and assigns
an action-aware target proportion of \textsc{SEEK} states (Table~\ref{tab:frg_seek_proportions}).
For a turn, the proportion depends on the total rotation of its maximal
contiguous run of turns in the same direction, rather than on the individual
turn angle. For example, two consecutive $15^\circ$ right turns form a
$30^\circ$ run, so each receives a target proportion of $50\%$.

\begin{table}[t]
\centering
\caption{Target SEEK proportions by expert action. For turns, the angle
denotes the cumulative rotation of consecutive turns in the same direction.}
\label{tab:frg_seek_proportions}
\begin{tabular}{lc}
\toprule
Expert action & SEEK proportion \\
\midrule
Move forward & $30\%$ \\
Turn left/right ($15^\circ$) & $30\%$ \\
Turn left/right ($30^\circ$) & $50\%$ \\
Turn left/right ($45^\circ$) & $75\%$ \\
Turn left/right (other angles) & $100\%$ \\
Stop & $0\%$ \\
\bottomrule
\end{tabular}
\end{table}

We then construct a deterministic mode plan. Decisions are grouped by action
detail: forward distance for \textsc{MoveForward}, or direction and individual
turn angle for turns. Each group initially receives the integer part of its
expected \textsc{SEEK} count. Remaining quotas are assigned to groups with
the largest fractional remainders until the total matches the rounded
expected count across all groups. Within each group, cumulative allocation
selects the required number of \textsc{SEEK} decisions; the rest are labeled
\textsc{NAV}. This preserves the action-conditioned proportions without
random sampling.

\subsection{Progress Reasoning Annotation}
\appendixsubsectionentry{Progress Reasoning Annotation}
\label{app:hrg:progress}
Progress labels are generated only for planned \textsc{SEEK} states. The VLM annotator receives the instruction, the provided subgoal list, the stride-two history, and the current front observation. It identifies the completed instruction prefix and the first unfinished subgoal, then refines them into the fields \texttt{completed\_tasks} and \texttt{next\_task}. The validation step requires the completed field to be a contiguous prefix of the instruction and the next-task field to be the subsequent unfinished subgoal.

\subsection{Evidence Seeking Annotation}
\appendixsubsectionentry{Evidence Seeking Annotation}
\label{app:hrg:evidence}
Evidence is annotated separately from progress. The annotator receives the instruction, history, and an action-conditioned subset of current replay views: the front view for forward actions and $15^\circ$ turns, and the front view plus the corresponding side view for $30^\circ$ and $45^\circ$ turns. It outputs a single concise \texttt{key\_evidence} statement.

\subsection{Dataset Construction}
\appendixsubsectionentry{Dataset Construction}
\label{app:hrg:dataset}

We serialize each expert decision as one training example. A \textsc{NAV}
example pairs the instruction and visual context with the expert navigation
action. A \textsc{SEEK} example additionally includes the evidence-seeking
interaction and a \texttt{<think>} block containing
\texttt{completed\_tasks}, \texttt{next\_task}, and
\texttt{key\_evidence}, followed by the expert navigation action. The
resulting 4K FRG prior dataset contains 111,141 decision-level examples
from 4,162 R2R-CE episodes: 68,795 \textsc{NAV} examples and
42,346 \textsc{SEEK} examples.

\paragraph{Seek mode example.}
\label{app:hrg:example}
The following record illustrates a \textsc{SEEK} training example.
Image placeholders denote attached images.
\begin{promptbox}
\begin{Verbatim}[commandchars=\\\[\],breaklines=true,breakanywhere=true,breaksymbolleft={},fontfamily=ptm,fontshape=it,fontsize=\small,formatcom={\fontencoding{T1}\selectfont}]
Instruction:
Go around the right side of the center unit and stop by the right-side doorway
with the dining table and mirror.

Initial mode:
SEEK (assigned by the macro-action quota plan)

Progress annotation input:
historical observations: <image> ... <image>
current front observation: <image>

Evidence annotation input:
front observation: <image>   right observation: <image>

Validated annotations:
{"completed_tasks": "None.",
 "next_task": "Go around the right side of the center unit.",
 "key_evidence": "A dining area and large framed wall decor are visible on the right."}

Final serialized training sample:
assistant: <seek>
environment: left view:<image> forward view:<image> right view:<image></seek>
assistant: <think>Completed tasks: None. Next task: Go around the right side of the
center unit. Key Evidence: A dining area and large framed wall decor are visible
on the right.</think><nav>The next action is turn right 45 degree.
\end{Verbatim}
\end{promptbox}

\subsection{Annotation Prompts}
\appendixsubsectionentry{Annotation Prompts}
\label{app:hrg:prompts}
\subsubsection{Progress Reasoning Stage-1 Prompt}
The following is the first prompt used for progress annotation, with instance-specific fields replaced by placeholders. Historical images are attached in chronological order before the current front observation.
\begin{promptbox}
\begin{Verbatim}[breaklines=true,breakanywhere=true,breaksymbolleft={},fontfamily=ptm,fontshape=it,fontsize=\small,formatcom={\fontencoding{T1}\selectfont}]
You are a vision-language navigation progress analyzer.
Determine the agent's semantic progress at the CURRENT observation.

Input:
- Historical observations are sampled every 2 primitive steps before the current
  observation and are provided in chronological order.
- The current observation is provided after the historical observations and is
  the only observation to label.
- The provided sub-goal list defines the only valid instruction boundaries.
- <HISTORY_COUNT> historical frame(s) will be attached before the current observation.

Navigation instruction:
<INSTRUCTION>

Provided sub-goals:
<SUBGOAL_LIST>

Task:
1. Compare the instruction with the visual evidence from the historical and
   current observations.
2. Determine which provided sub-goals form a fully completed contiguous prefix.
3. Determine the first unfinished provided sub-goal after that completed prefix.

Progress rules:
- Completed tasks must be None or the exact text of the completed contiguous
  prefix from the provided sub-goal list.
- Next task must be the first provided sub-goal not included in Completed tasks.
- If all provided sub-goals are completed, Next task must be Stop.
- Do not split a provided sub-goal further, even if it contains multiple actions.
- Do not skip an earlier unfinished sub-goal.
- Use the original wording from the provided sub-goal list whenever possible.
- Partial completion does not count as completion.

Output format (MUST follow exactly):
<Analysis>
- Briefly state only the instruction-relevant historical/current visual evidence.
- State which provided sub-goals are fully completed, if any.
- State why the next sub-goal is the first unfinished one.
</Analysis>
<Answer>
Completed tasks: <None or the exact completed sub-goal prefix>
Next task: <the first unfinished sub-goal or Stop>
</Answer>

Important constraints:
- No markdown headings.
- No extra text outside the tags.
\end{Verbatim}
\end{promptbox}

\subsubsection{Progress Reasoning Stage-2 Prompt}
The second prompt refines the stage-1 response into the canonical progress fields used for serialization.
\begin{promptbox}
\begin{Verbatim}[breaklines=true,breakanywhere=true,breaksymbolleft={},fontfamily=ptm,fontshape=it,fontsize=\small,formatcom={\fontencoding{T1}\selectfont}]
You are an expert navigation progress summarizer.
You will receive a navigation instruction and a previous model output describing
the agent's progress from the historical observations to the current observation.
Your task is to refine that previous output into two structured labels aligned
with the navigation instruction.

Navigation instruction:
<INSTRUCTION>

Previous model analysis:
<STAGE1_ANALYSIS>

Previous model answer:
Completed tasks: <STAGE1_COMPLETED_TASKS>
Next task: <STAGE1_NEXT_TASK>

Rules:
- Treat the previous model analysis and answer only as evidence to be checked
  against the navigation instruction.
- Use the navigation instruction as the only textual task reference.
- Preserve only progress that is both visually supported by the previous model
  output and consistent with the navigation instruction.
- Do not add new actions, landmarks, rooms, objects, or interpretations.
- Do not split the navigation instruction into new sub-goals.
- Do not merge unrelated instruction clauses.
- Do not repeat the visual analysis or describe observations in the final output.
- Preserve the original wording of the navigation instruction whenever possible.
- Completed tasks must be None or the completed instruction clauses already
  supported by the previous model output, starting from the beginning of the instruction.
- Next task must be the first unfinished instruction clause after the completed prefix.
- Do not treat a partially completed instruction clause as completed.
- Do not infer completion from a future action or from a landmark merely being visible.
- If no instruction clause is complete, use "None" for completed_tasks.
- If all instruction steps are complete, use "Stop" for next_task.
- If the previous model output is ambiguous, incomplete, or conflicts with the
  navigation instruction, choose the most conservative completed prefix.
- Do not output sub-goal indexes, explanations, Markdown, or any extra fields.

Output JSON only, without Markdown fences:
{
  "completed_tasks": "None or the completed instruction prefix",
  "next_task": "The first unfinished instruction clause or Stop"
}
\end{Verbatim}
\end{promptbox}

\Needspace{8\baselineskip}
\subsubsection{Evidence Seeking Annotation Prompt}
The following prompt is used for action-conditioned evidence annotation. The view names and the final directional phrase are instantiated from the target macro action.
\begin{promptbox}
\begin{Verbatim}[breaklines=true,breakanywhere=true,breaksymbolleft={},fontfamily=ptm,fontshape=it,fontsize=\small,formatcom={\fontencoding{T1}\selectfont}]
Imagine you are a robot programmed for navigation tasks. We provide you with
the historical observations, the current views and navigation instruction
listed below.

Input:

Navigation instruction:
<INSTRUCTION>

Historical observations:
<HISTORY>

Current views:
<ACTION_CONDITIONED_VIEWS>

Each entry is a view name followed by its corresponding image. Only the views
listed above are provided for this sample.

Task:
You need to describe concise, task-relevant visual evidence based on the
navigation instruction and current view. The historical observations are
provided to indicate the task progress to help you describe more reliably.
<DIRECTION_QUESTION>

Evidence rules:
- Describe only the visible evidence that is relevant to the navigation instruction.
- Focus on landmarks, objects, doorways, corridors, rooms, targets, or spatial
  relations that help interpret the instruction.
- Do not describe the entire scene.
- Do not predict the next action.
- Do not infer objects or locations that are not visibly supported.
- Keep the evidence concise: describe at most one or two instruction-relevant
  visual elements.
- Output exactly one line in the form "Reasoning: <brief reasoning>. Key Evidence:
  I see <one concise instruction-relevant visual description> <DIRECTION>.".
- The final Key Evidence segment must end exactly with <DIRECTION>.

Response strictly in this format:
Reasoning: <brief reasoning>. Key Evidence: I see <one concise
instruction-relevant visual description> <DIRECTION>.

If no clear instruction-relevant evidence is visible, output:
Reasoning: <brief reasoning>. Key Evidence: I see no clear instruction-relevant
evidence <DIRECTION>.
\end{Verbatim}
\end{promptbox}

These annotations are serialized into $\mathcal{D}_{\mathrm{nav}}$ and $\mathcal{D}_{\mathrm{seek}}$ following Eq.~\ref{eq:prior_dataset}, and the resulting $\mathcal{D}_{\mathrm{prior}}$ is used for the supervised cold-start training described in Appendix~\ref{app:cold_start}.

\raggedbottom
\section{Training Details}
\appendixsectionentry{Training Details}
\label{app:training_details}

\subsection{Supervised Fine-Tuning (FRG-SFT)}
\appendixsubsectionentry{Supervised Fine-Tuning (SFT)}
\label{app:cold_start}

We initialize SeekVLN from the pretrained Aux-Think~\citep{auxthink} model by supervised fine-tuning on $\mathcal{D}_{\mathrm{prior}}$. Cold-start training learns the restricted first-token decision between \texttt{<nav>} and \texttt{<seek>}, while also learning the direct navigation response or, after evidence acquisition, the structured progress reasoning and navigation response.

\paragraph{Objective.}
We optimize a restricted mode loss together with a weighted response loss:
\begin{equation}
\mathcal{L}_{\mathrm{SFT}}=
\lambda_{\mathrm{mode}}\mathcal{L}_{\mathrm{mode}}+
\mathcal{L}_{\mathrm{resp}}.
\end{equation}
For the target mode $m_t^*\in\{\texttt{<nav>},\texttt{<seek>}\}$, the mode loss is
\begin{equation}
\mathcal{L}_{\mathrm{mode}}=
-\frac{1}{|\mathcal{D}_{\mathrm{prior}}|}
\sum_t
\log\frac{\exp z_{t,m_t^*}}
{\exp z_{t,\texttt{<nav>}}+\exp z_{t,\texttt{<seek>}}},
\end{equation}
where $z_{t,\cdot}$ are the logits at the first assistant generation position. This restricted normalization prevents the rare mode decision from being diluted by the much longer free-form response.

For response tokens, we use
\begin{equation}
\mathcal{L}_{\mathrm{resp}}=
-\frac{1}{N_{\mathrm{valid}}}
\sum_t\sum_i \mu_{t,i}\omega_{t,i}
\log\pi_\theta(y_{t,i}\mid x_t,y_{t,<i}),
\end{equation}
Here, $\mu_{t,i}$ selects supervised tokens and $\omega_{t,i}$ upweights structural tokens. The environment content strictly between \texttt{<seek>} and \texttt{</seek>} is masked. During interaction, the environment prefills \texttt{</seek>} after the supplementary views, and the policy generates \texttt{<think>} to begin reasoning. Both \texttt{</seek>} and \texttt{<think>} remain supervised during SFT.

The cold-start training configuration is summarized in
Table~\ref{tab:cold_start_config}.

\begin{table}[H]
\centering
\small
\caption{FRG SFT training configuration.}
\label{tab:cold_start_config}
\begin{tabular}{p{0.34\linewidth}p{0.58\linewidth}}
\toprule
Component & Setting \\
\midrule
Base model & Pretrained Aux-Think~\citep{auxthink} \\
Trainable modules & Language model and multimodal projector; visual encoder frozen \\
Maximum sequence length & 512 tokens \\
Learning rate & $2\times10^{-5}$ \\
Special tokens & \texttt{<nav>}, \texttt{<seek>}, \texttt{</seek>},
\texttt{<think>}, \texttt{</think>} \\
Observations input & Current observation and up to eight sampled historical frames;
left, front, and right views are added in \textsc{SEEK} mode \\
\bottomrule
\end{tabular}
\end{table}

% =====================================================================
% Appendix: Reinforcement Learning Details
% Requires: \usepackage{booktabs}, \usepackage{amsmath}
% If the document is two-column, change {table} to {table*}.
% =====================================================================

\subsection{Reinforcement Fine-Tuning (C2PO)}
\appendixsubsectionentry{Reinforcement Fine-Tuning (C2PO)}
\label{app:rl_details}

Starting from the FRG-SFT checkpoint, we optimize SeekVLN with
decision-level PPO~\citep{schulman2017proximalpolicyoptimizationalgorithms}:
the complete response at each navigation decision is treated as one policy
action. The environment reward combines the counterfactual and adaptive
outcome terms defined in Sec.~3.3, with no per-decision action penalty.
We additionally use adaptive KL regularization against the frozen SFT
reference policy. Training settings are summarized in
Table~\ref{tab:rl_hyperparams}.

\begingroup
\small
\begin{longtable}{@{}p{0.39\linewidth}p{0.56\linewidth}@{}}
\caption{C2PO RFT training configuration.}
\label{tab:rl_hyperparams}\\
\toprule
Component & Setting \\
\midrule
\endfirsthead
\caption[]{C2PO RFT training configuration (continued).}\\
\toprule
Component & Setting \\
\midrule
\endhead
\bottomrule
\endfoot
\multicolumn{2}{@{}l}{\textit{Model and data}} \\*
Initialization & SeekVLN-FRG-SFT \\
Trainable modules & Language model and multimodal projector; visual encoder frozen \\
Value function & Separate model initialized from the SFT checkpoint \\
Visual context & Nine frames per decision \\
Training data & 640 episodes sampled from the R2R-CE train split \\
\midrule
\multicolumn{2}{@{}l}{\textit{Rollout and PPO}} \\*
Episode workers & 16, asynchronous \\
Episodes per update & 32 \\
Total updates & 20 \\
Advantage estimation & Decision-level GAE; $\gamma=1.0$, $\lambda=1.0$ \\
Advantage normalization & Per-update RMS, floor $1.0$ \\
PPO clip ratio / epochs & $0.2$ / $1$ \\
Actor / critic learning rate & $4\times10^{-6}$ / $1\times10^{-5}$ \\
Mini-batch size & 128 decisions \\
Micro-batch per GPU & 4 decisions \\
\multicolumn{2}{@{}l}{\textit{Reward and regularization}} \\*
Counterfactual reward & Applied only at \textsc{SEEK} decisions; $w$ times a contrast clipped to $[-0.2,0.2]$ \\*
Adaptive outcome reward & $1+0.2\,\mathrm{SPL}$ on success; $-0.5$ on failure \\*
KL reference & Frozen FRG-SFT policy \\*
Initial KL coefficient / target & $0.2$ / $0.3$ per response \\*
KL update horizon & $10^4$ \\*
\midrule
Hardware & 8 NVIDIA RTX 6000D GPUs, FSDP2 \\
\end{longtable}
\endgroup

\flushbottom
\section{Ablation Study}
\appendixsectionentry{Ablation Study}
\label{sec:ablation}

We ablate the two training stages separately: Future-guided Reverse Generation (FRG), which supplies progress-reasoning and evidence-seeking supervision, and Counterfactual Contrastive Policy Optimization (C2PO), which optimizes the policy through counterfactual branch comparison. Within each comparison, variants share the same backbone, training budget, and optimization settings except for the component under study.

We evaluate all ablations on a trajectory-deduplicated subset of R2R-CE Val-Unseen: one instruction is retained for each of the 613 unique expert trajectories. The subset preserves all 11 unseen Matterport3D scenes while avoiding repeated evaluation of the same route under multiple instructions.

\subsection{Analysis of FRG}
\appendixsubsectionentry{Analysis of FRG}
\label{sec:hrg_ablation}

FRG provides two complementary supervision signals: progress reasoning tracks completed and upcoming instruction subgoals, while evidence-seeking examples associate trigger decisions with task-relevant visual evidence. We remove each signal independently to assess its contribution.

\begin{table}[H]
\centering
\begin{minipage}[t]{0.50\textwidth}
\centering
\captionof{table}{FRG ablations on the R2R-CE Val-Unseen-613 subset.}
\label{tab:hrg_ablation}
\resizebox{\linewidth}{!}{
\begin{tabular}{lccc}
\toprule
& \multicolumn{3}{c}{R2R-CE Val-Unseen-613} \\
\cmidrule(lr){2-4}
& NE $\downarrow$ & SR $\uparrow$ & SPL $\uparrow$ \\
\midrule

Aux-Think (base model)
& 5.80 & 52.0 & 45.0 \\

w/o Evidence Seeking
& \underline{4.86} & 58.6 & \underline{53.3} \\

w/o Progress Reasoning
& 4.99 & \underline{58.7} & 53.0 \\

\rowcolor{blue!15}
\textbf{Full FRG}
& \textbf{4.84} & \textbf{60.7} & \textbf{55.8} \\
\bottomrule
\end{tabular}
}
\end{minipage}\hfill
\begin{minipage}[t]{0.48\textwidth}
\centering
\captionof{table}{C2PO ablations on the R2R-CE Val-Unseen-613 subset.}
\label{tab:c2po_reward}
{\setlength{\tabcolsep}{2.5pt}
\renewcommand{\arraystretch}{1.28}
\resizebox{\linewidth}{!}{
\begin{tabular}{@{}lcccc@{}}
\toprule
& \multicolumn{4}{c}{R2R-CE Val-Unseen-613} \\
\cmidrule(lr){2-5}
& NE $\downarrow$ & SR $\uparrow$ & SPL $\uparrow$ & Seek Rate (\%) \\
\midrule

SeekVLN-FRG-SFT
& 4.84 & 60.7 & 55.8 & 15.9 \\

w/o CF Reward
& \underline{3.94} & \underline{65.1} & \underline{60.4} & 34.3 \\

\rowcolor{blue!15}
\textbf{Full C2PO}
& \textbf{3.66} & \textbf{68.5} & \textbf{62.4} & 29.3 \\

\bottomrule
\end{tabular}
}
}
\end{minipage}
\end{table}

Table~\ref{tab:hrg_ablation} shows that removing progress reasoning lowers SR by 1.96 points and SPL by 2.84 points; removing evidence-seeking supervision lowers them by 2.13 and 2.54 points, respectively. NE also rises in both variants. The consistent degradation indicates that both signals contribute to navigation performance.

\subsection{Analysis of C2PO}
\appendixsubsectionentry{Analysis of C2PO}
\label{sec:c2po_ablation}
To isolate the contribution of branch-level credit assignment in C2PO, we remove the counterfactual reward while retaining the FRG-SFT initialization, adaptive outcome reward, and all other reinforcement fine-tuning settings.

As shown in Table~\ref{tab:c2po_reward}, adding the counterfactual reward raises SR from 65.09\% to 68.52\% and SPL from 60.40\% to 62.36\%, while reducing NE from 3.94\,m to 3.66\,m. The seek rate also falls from 34.3\% to 29.3\%. Together, these results suggest that counterfactual credit assignment improves navigation through more selective and effective evidence-seeking decisions.

\begin{figure}[H]
    \centering
    \includegraphics[width=0.96\linewidth]{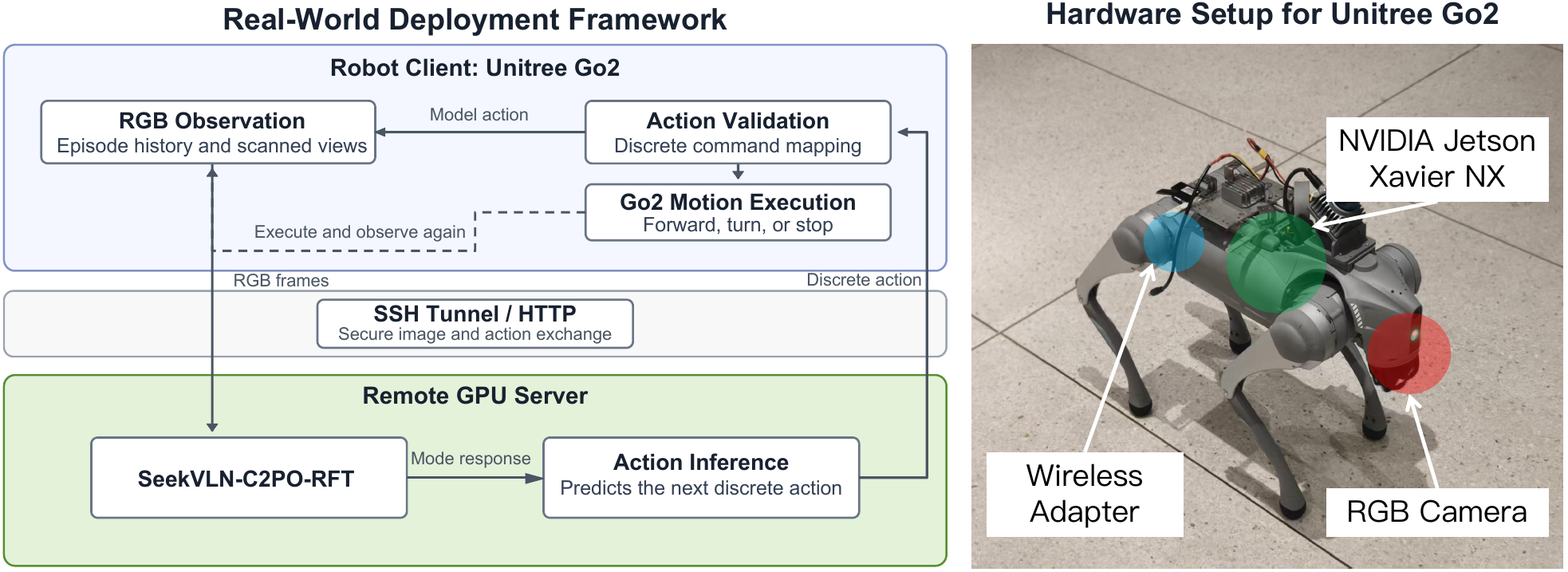}
    \caption{Real-world SeekVLN deployment. The Go2 robot handles sensing and
    motion, while a remote GPU server performs policy inference. The hardware
    setup is shown on the right.}
    \label{fig:seekvln-deployment}
\end{figure}

\section{Real-World Deployment}
\appendixsectionentry{Real-World Deployment}
\label{app:real-world-deployment}

We deploy SeekVLN-C2PO-RFT on a Unitree Go2 using the client--server architecture in
Fig.~\ref{fig:seekvln-deployment}. The robot handles RGB observation, view
acquisition, and motion execution, while the policy runs on a remote GPU
server. An SSH tunnel and HTTP interface carry image observations to the
server and return discrete actions to the robot.

At each decision, the model selects either a navigation or visual-search mode.
For visual search, the robot acquires left, forward, and right views by turning
in place. The model output is mapped to a finite set of Go2 commands: forward
motion in 25 cm increments, discrete left or right turns, or stop. The local
controller validates the output and executes one command before acquiring the
next observation. A stop prediction indicates model-declared completion; it
does not independently verify that the goal has been reached.

Camera images are converted to RGB and resized to $448\times448$ pixels for
the vision encoder. Up to nine real observations are provided as history;
shorter histories are not padded. For reproducibility, each run records the
source images, model inputs and outputs, robot states, actions, and timing
information. Motion commands are bounded, and the robot stops on inference,
communication, observation, or control failure.

\end{document}